\documentclass[runningheads]{llncs}
\pdfoutput=1
\usepackage{graphicx}
\usepackage{amsmath,amssymb} 
\usepackage{color}
\usepackage[width=122mm,left=12mm,paperwidth=146mm,height=193mm,top=12mm,paperheight=217mm]{geometry}
\begin{document}

\title{Statistics of RGBD Images}

\titlerunning{Statistics of RGBD Images}

\authorrunning{Rosenbaum and Weiss}

\author{Dan Rosenbaum and Yair Weiss}
\institute{School of Computer Science and Engineering\\
Hebrew University of Jerusalem}

\maketitle

\begin{abstract}

Cameras that can measure the depth of each pixel in addition to its
color have become easily available and are used in many consumer
products worldwide.  Often the depth channel is captured at lower
quality compared to the RGB channels and different algorithms have
been proposed to improve the quality of the D channel given the RGB
channels. Typically these approaches work by assuming that edges in
RGB are correlated with edges in D. 

In this paper we approach this problem from the standpoint of natural
image statistics.  We obtain examples of high quality RGBD images from
a computer graphics generated movie (MPI-Sintel) and we use these
examples to compare different probabilistic generative models of RGBD
image patches. We then use the generative models together with a
degradation model and obtain a Bayes Least Squares (BLS) estimator of
the D channel given the RGB channels. Our results show that learned
generative models outperform the state-of-the-art in improving the
quality of depth channels given the color channels in natural images
even when training is performed on artificially generated images.
 
\end{abstract}

\section{Introduction}
\begin{figure}
\centerline{
\includegraphics[width=0.5\columnwidth]{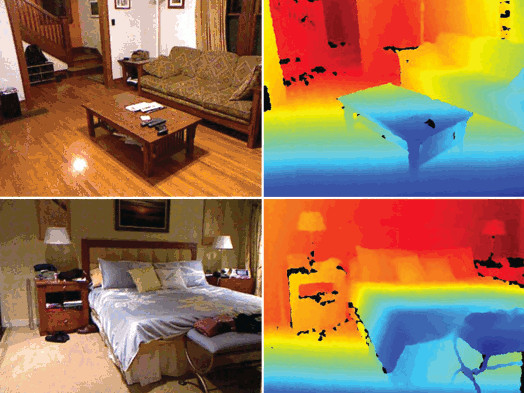}}
\caption[]{Examples of RGBD images from the NYU Depth V2
  datatset. The depth channel often contains missing values and the
  depth is typically of lower resolution and more noisy than the
  RGB. In this paper we approach the problem of improving the D
  channel given RGB using natural image statistics.}
\label{nyu-fig}
\end{figure}

Figure~\ref{nyu-fig} shows examples from the NYU Depth V2 dataset~\cite{Silberman:ECCV12}. Each
scene is captured with a Kinect sensor and a color image is available
along with a depth image. Ten years ago it may have been hard to
believe that a depth image of such quality will be attainable with a
sensor that costs less than 200 dollars, but today RGBD cameras are
ubiquitous and have enabled a large suite of consumer
applications. Despite the impressive improvement in RGBD technology,
the quality of the depth channel is still lacking. As can be seen in
the figure, the depth channel often has missing pixels. Many of these
missing pixels occur at object discontinuities where the different
sensors used to measure depth have a viewpoint disparity. Others occur
at specular objects. In addition,  the depth image is often noisy and
at a poorer resolution compared to the RGB channels. 

In recent years, several authors have proposed improving the quality
of the D channel based on the RGB channel~\cite{liu2012guided,liu2013kinect}. The vast majority of these
approaches are based on assuming that depth edges are more likely to
occur at intensity edges and this leads to a natural use of the joint
bilateral filter~\cite{qi2013structure,richardt2012coherent}. Silverman and Fergus~\cite{Silberman:ECCV12} used
the colorization by optimization framework of Levin et
al.~\cite{levin2004colorization} to obtain a weighted least squares problem for
filling in missing pixels where the weights are based on the
assumption that neighboring pixels with similar colors should have
similar depths. 

As pointed out by Lu et al.~\cite{Lu2014depth}, the assumption of correlation between
color edges and depth edges may be insufficient to improve the quality
of the depth image. In particular, they pointed out that both the
color and the depth image are often subject to noise and that previous
approaches did not handle this noise well. They suggested a
statistical model of RGBD patches which is based on the assumption
that similar patches in the image define a low rank matrix. Their approach
outperformed approaches such as joint bilateral filtering, even when
the color image was first denoised using a denoising algorithm. 

In this paper we approach the problem of RGBD restoration from the
standpoint of natural image statistics.  We are motivated by the success of learning based methods that achieve excelllent performance in image restoration~\cite{zoran2011learning,schmidt2014shrinkage,burger2012image} by learning from a large database of clean images. In the case of RGBD the challenge is to obtain clean examples and we take advantage of a  computer graphics generated movie
(MPI-Sintel~\cite{Butler:ECCV:2012})  for this task. We use the clean examples to compare existing approaches and to learn new generative models for the patches. We then use the
generative models together with a degradation model and obtain a Bayes
Least Squares (BLS) estimator of the D channel given the RGB
channels. Our results show that learned generative models outperform the
state-of-the-art in improving the quality of depth channels given the
color channels in natural images even when training is performed on
artificially generated images.

\section{Density models for depth}
All methods for depth enhancement incorporate some assumption about the depth itself and sometimes about its dependence on the color channels. 
Typical assumptions are that the depth is usually smooth and that depth boundaries are correlated to color boundaries.

One way to compare different assumptions is to formulate them as density models for depth. 
Instead of using depth values in meters, we use the common representation of $1/depth$ or \emph{disparity}. This has the advantage that background pixels with depth infinity which are very common translate to a mode in zero, and the precision is higher for closer objects.

We will evaluate the following density models, where $d$ is a vector of disparity pixels:
\subsubsection*{DL2} The smoothness is modeled by giving a quadratic penalty to the spatial derivatives of disparity:
$$J(d) = \sum_p d_x(p)^2 + d_y(p)^2$$
where $d_x(p)$ and $d_y(p)$ are the $x$ and $y$ derivatives of disparity at pixel $p$.
This can be formulated as a multivariate Gaussian over the disparity using a matrix $A$ that takes all the derivatives of $d$. To make the covariance positive definite we add the indentity matrix times a small constant. 
\begin{equation}
Pr(d) = \frac{1}{Z} e^{-\lambda \sum_p d_x(p)^2 + d_y(p)^2} \approx \frac{1}{Z}e^{- d^\top (\lambda A^\top A + \epsilon I)d}
\end{equation} 

\subsubsection*{DL1} The smootheness is modeled by giving an absolute value penalty to the spatial derivatives of disparity:
$$J(d) = \sum_p |d_x(p)| + |d_y(p)|$$
This can be formulated as a multivariate Laplacian over $d$ using the same derivative matrix $A$ as above:
\begin{equation}
Pr(d) = \frac{1}{Z} e^{-\lambda \sum_p |dx(p)| + |dy(p)|} \approx \frac{1}{Z}e^{- \|(\lambda A + \epsilon I)d\|_1}
\end{equation} 
Here the normalization cannot be computed in closed form, making this model hard to use for measuring likelihood. 

\subsubsection*{DL2$\vert$int} Here we use a weighted quadratic penalty on the derivatives of disparity, where the weights $w(p)$ depend on the color image:
$$J(d) = \sum_p w_x(p) ~d_x(p)^2 + w_y(p) ~d_y(p)^2$$
In order to encourage disparity edges to correlate with color edges, the weights are computed as a function of the color derivatives in the same location $c_x(p)$ and $c_y(p)$ as following: 
$$w_x(p) = e^{-\frac{1}{\sigma^2} c_x(p)^2} ~~~ w_y(p) = e^{-\frac{1}{\sigma^2} c_y(p)^2}$$
giving derivatives that cross color edges a lower weight. This is the model of the colorization by optimization code~\cite{levin2004colorization} used in \cite{Silberman:ECCV12}.

The model can be formulated as a conditional multivariate Gaussian over $d$ using the same derivative matrix $A$ and an additional diagonal weight matrix that depends on the color $W(c)$:
\begin{equation}
Pr(d|c) = \frac{1}{Z} e^{-\lambda \sum_p w_x(p) d_x(p)^2 + w_y(p) d_y(p)^2} \approx \frac{1}{Z}e^{- d^\top (\lambda A^\top W(c)  A + \epsilon I)d}
\end{equation} 
For simplicity, and since we haven't noticed any significant difference, we reduce the RGB channels to a single intensity channel.

\subsection{Evaluation of density models}
\begin{figure}
\includegraphics[width=0.32\textwidth]{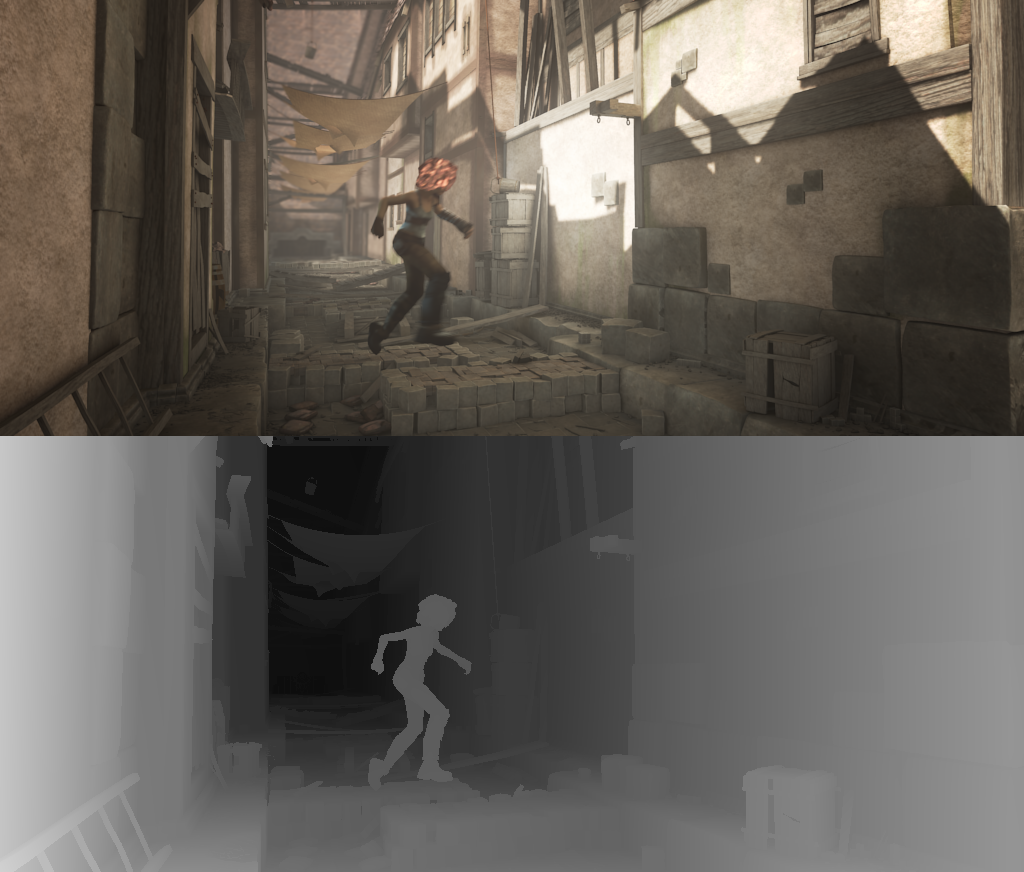} 
\includegraphics[width=0.32\textwidth]{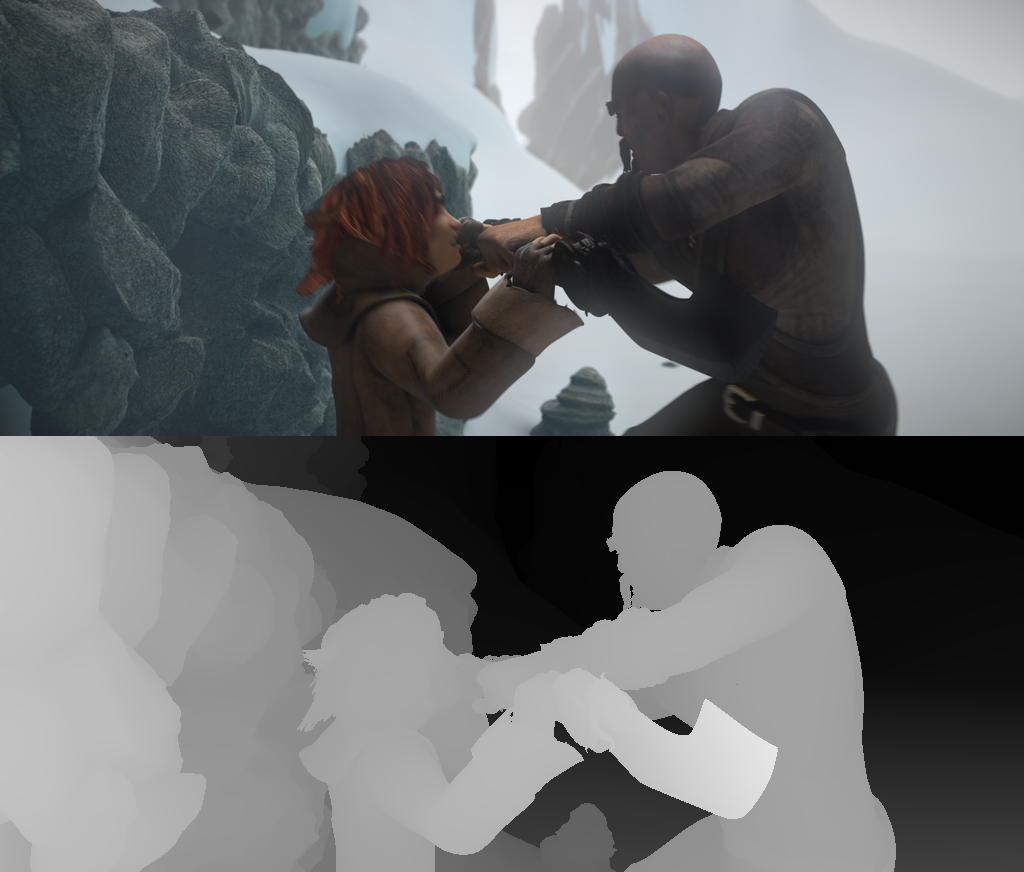} 
\includegraphics[width=0.32\textwidth]{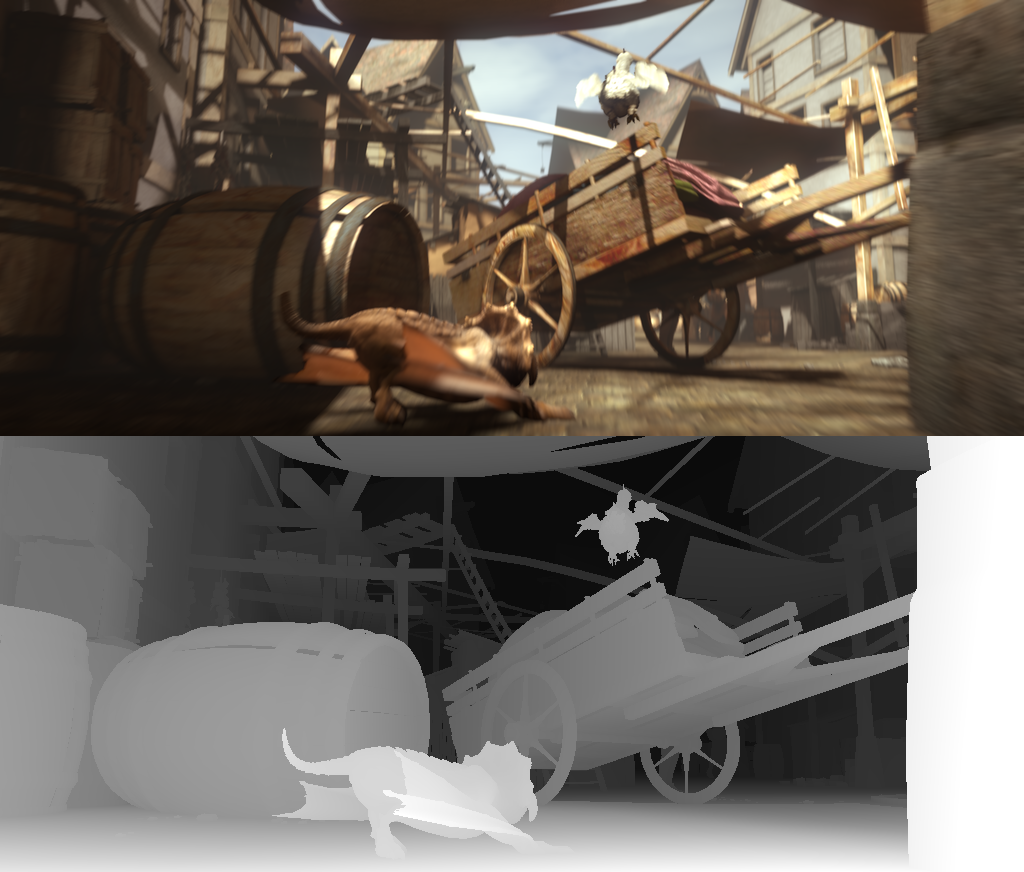} 
\caption{The Sintel dataset. Top: color images. Bottom: disparity=1/depth images. Using high quality depth images allows us to evaluate and learn density models.}
\label{fig:sintel}
\end{figure}

The challenge in applying learning techniques to RGBD data is to obtain a large dataset of clean images. Previous works  (e.g.~\cite{HuangLM00}) used the output of a depth sensor in order to estimate the statistics but these statistics themselves may already be corupted. Here we use a highly realistic computer graphics generated dataset, the MPI-Sintel  dataset~\cite{Butler:ECCV:2012} (figure \ref{fig:sintel}). We divided the 23 scenes of Sintel to 16 training set scenes and 7 test set scenes.
We follow roughly the approach of Rosenbaum and Weiss~\cite{rosenbaum2013learning}
and use the training set to tune the parameters $\lambda$ and $\epsilon$ for each model and we use the test set to evaluate the different models.

\begin{figure}
\centering
\includegraphics[width=1\textwidth]{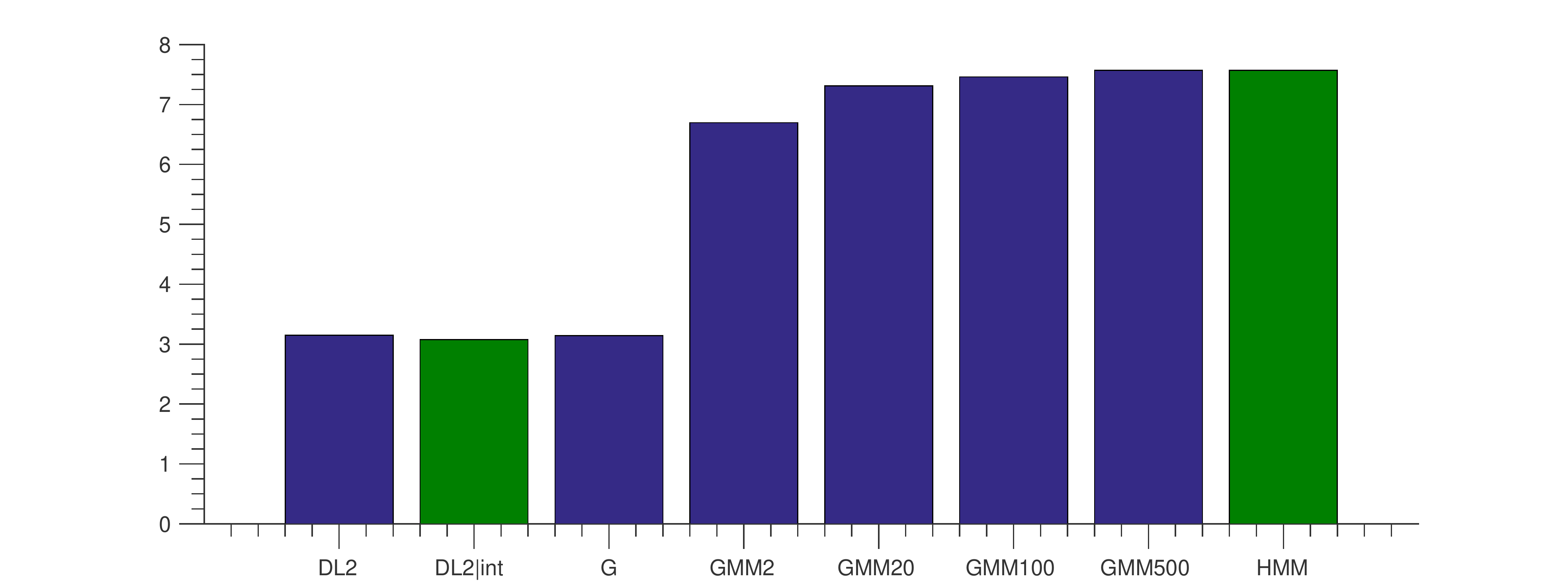} 
\caption{The log-Likelihood of hand-crafted density models and learned density models of disparity. A GMM model with enough components outperforms other models. Models that are conditioned on the intensity (shown in green) have a very similar log-likelihood to the unconditional models.}
\label{fig:logl}
\end{figure}

\subsubsection{Likelihood}
The first way to evaluate the density models is by the likelihood on the test set. 
Since all density models need to integrate to 1 over all possible values, models that give high likelihood to a set of ground truth disparity images are models that capture frequent properties of the data.
Figure \ref{fig:logl} shows the resulting log-likelihood per pixel for the different models. We can see that the log-likelihood for DL2 and DL2$\vert$int are very similar. Since we can't compute exactly the noramlization contstant of DL1 we don't use it here.

\begin{figure}
\centering
\renewcommand{\tabcolsep}{1.5pt}
\begin{tabular}{cccccc}
GT & DL2 & G & GMM2 & GMM20 & GMM500  \\
\includegraphics[width=0.155\textwidth]{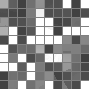} &
\includegraphics[width=0.155\textwidth]{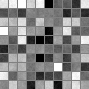} &
\includegraphics[width=0.155\textwidth]{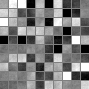} &
\includegraphics[width=0.155\textwidth]{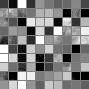} &
\includegraphics[width=0.155\textwidth]{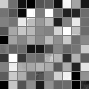} &
\includegraphics[width=0.155\textwidth]{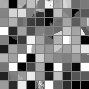} \\[0.3cm]
\includegraphics[width=0.155\textwidth]{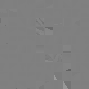} & 
\includegraphics[width=0.155\textwidth]{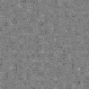} &
\includegraphics[width=0.155\textwidth]{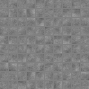} &
\includegraphics[width=0.155\textwidth]{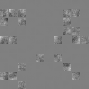} &
\includegraphics[width=0.155\textwidth]{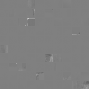} &
\includegraphics[width=0.155\textwidth]{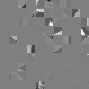} 
\end{tabular}
\caption{Patches from the ground truth (GT) vs. patches that were randomaly generated from different models. For better visibility, the bottom line shows the same patches with the DC substracted from each patch. Patches generated from a GMM with enough components exhibit similar properties as the ground truth: patches are usually very flat, and occasionally contain an edge.}
\label{fig:gsamps}
\end{figure}

\subsubsection{Patch generation}
A second way to evaluate the models is by using them to generate random data and testing for the visual similarity with ground truth data. We ommit DL1 from this test again since it does not allow for closed form generation of samples.
Figure \ref{fig:gsamps} shows ground truth $8 \times 8$ patches and patches generated from DL2. For better visibility we show all patches also with their DC (average value) subtracted. Looking at the ground truth disparity patches we can see that it is usually flat but occasionally contain a boundary edge. 
In comparison, patches generated from DL2 are a bit noisier and contain no structure.

\begin{figure}
  \centering
\renewcommand{\tabcolsep}{0.5pt}
\begin{tabular}{cc}
  \begin{tabular}{c}\small{intensity}\\\small{disparity}\end{tabular}
  \begin{tabular}{c}\includegraphics[trim={0 0 8cm 0},clip,width=0.9\textwidth]{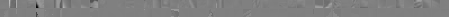}\end{tabular}
\end{tabular}
\caption{Ground truth patches of disparity together with the corresponding intensity patch (all patches are shown without the DC). The correlation between intensity and disparity is not very strong: Intensity edges can occur with no corresponding disparity edge (due to texture), and disparity edges can occur with no corresponding intensity edge (due to motion blur and atmospheric effects).}
\label{fig:intVsDisp}
\end{figure}

\begin{figure}
  \centering
  \renewcommand{\tabcolsep}{5pt}
  \begin{tabular}{lccc}
    intensity patch  &  \begin{tabular}{c}\includegraphics[width=0.05\textwidth]{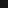}\end{tabular} &  \begin{tabular}{c}\includegraphics[width=0.05\textwidth]{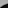}\end{tabular} &  \begin{tabular}{c}\includegraphics[width=0.05\textwidth]{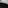}\end{tabular} \\
 DL2$\vert$int &  \begin{tabular}{c}\includegraphics[width=0.155\textwidth]{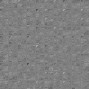}\end{tabular} &  \begin{tabular}{c}\includegraphics[width=0.155\textwidth]{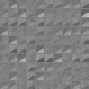}\end{tabular} &  \begin{tabular}{c}\includegraphics[width=0.155\textwidth]{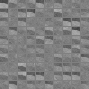}\end{tabular} \\
 HMM &  \begin{tabular}{c}\includegraphics[width=0.155\textwidth]{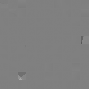}\end{tabular} &  \begin{tabular}{c}\includegraphics[width=0.155\textwidth]{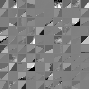}\end{tabular} &
 \begin{tabular}{c}\includegraphics[width=0.155\textwidth]{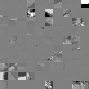}\end{tabular}\\
\end{tabular}
\caption{Disparity patches genereated conditionally given the intensity patches on the top. The DL2$\vert$int generates patches with edges that match exactly the intensity edge. The HMM can only approximate the edge form but can capture the distribution in its orientation and translation, and also the probability that the edge is missing.}
\label{fig:gsampsGivenInt}
\end{figure}

In figure \ref{fig:intVsDisp} we show the relationship between the disparity and intensity. The ground truth patches of disparity are shown together with the corresponding intensity patch. It can be seen that the relationship is not straightforward. First, in some cases both patches contain some structure which is not exactly correlated. Second, there are intensity edges without a corresponding disparity edge and there are disparity edge without a corresponding intensity edge. While the first direction can be attributed to many texture edges in intensity, the second direction which is perhaps more surprising is due to motion blur and atmospheric effects which are real effects that are deliberately modeled in the Sintel dataset\footnote{we use Sintel's \emph{final} pass of the intensity channel.}. 

Figure \ref{fig:gsampsGivenInt} shows patches generated from DL2$\vert$int given 3 different patches of intensity. The generated patches usually match the intensity patch exactly, and sometimes do not contain a visible structure. The advantage of the patches generated with DL2$\vert$int over patches of DL2 is evident since it alows for spatial structure that is very similar to the ground truth patches, however it is not clear whether the dependence on the intensity is modeled correctly.

\subsubsection{Patch restoration}
A third way to evaluate density models is to  use them in inference tasks and measure the quality of the results. 
Given ground truth patches we add noise using a known noise model and use Bayes Least Squares (BLS) to estimate the clean patches again. 
We measure the quality of the estimation using the $PSNR = 10 log_{10} (1/L)$, which is a function of the average squared loss over all restored patches:
$$L(\{\hat{d}\}) = \frac{1}{N}\sum_{i=1}^N \|\hat{d}_i-d_i\|_2^2$$ 
If the patches were generated from a known density model, then BLS inference with the true model would result in the optimal PSNR. 
Therefore we expect that BLS inference with models that are closer to the true density will result in a bigger PSNR.

\begin{figure}
\begin{tabular}{cc}
\begin{tabular}{c}$\sigma=5/255$\end{tabular} & 
\begin{tabular}{c}\includegraphics[width=0.8\textwidth]{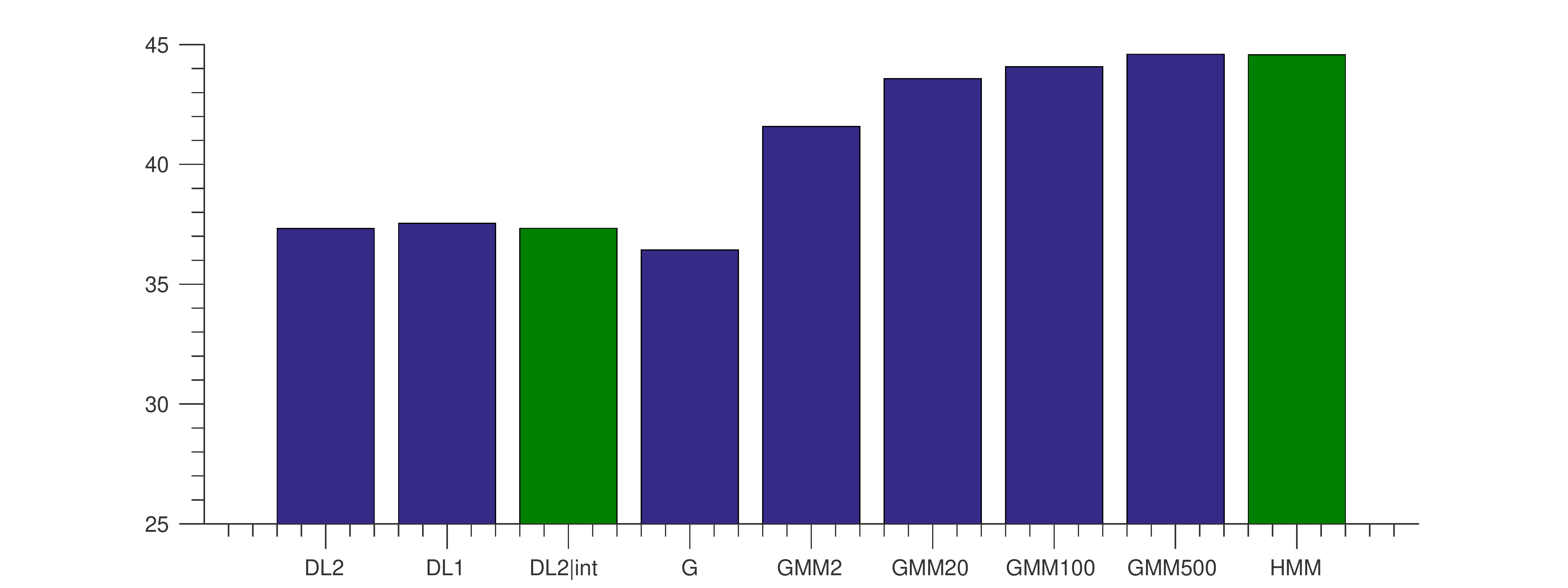}\end{tabular} \\
\begin{tabular}{c}$\sigma=15/255$\end{tabular} &
\begin{tabular}{c}\includegraphics[width=0.8\textwidth]{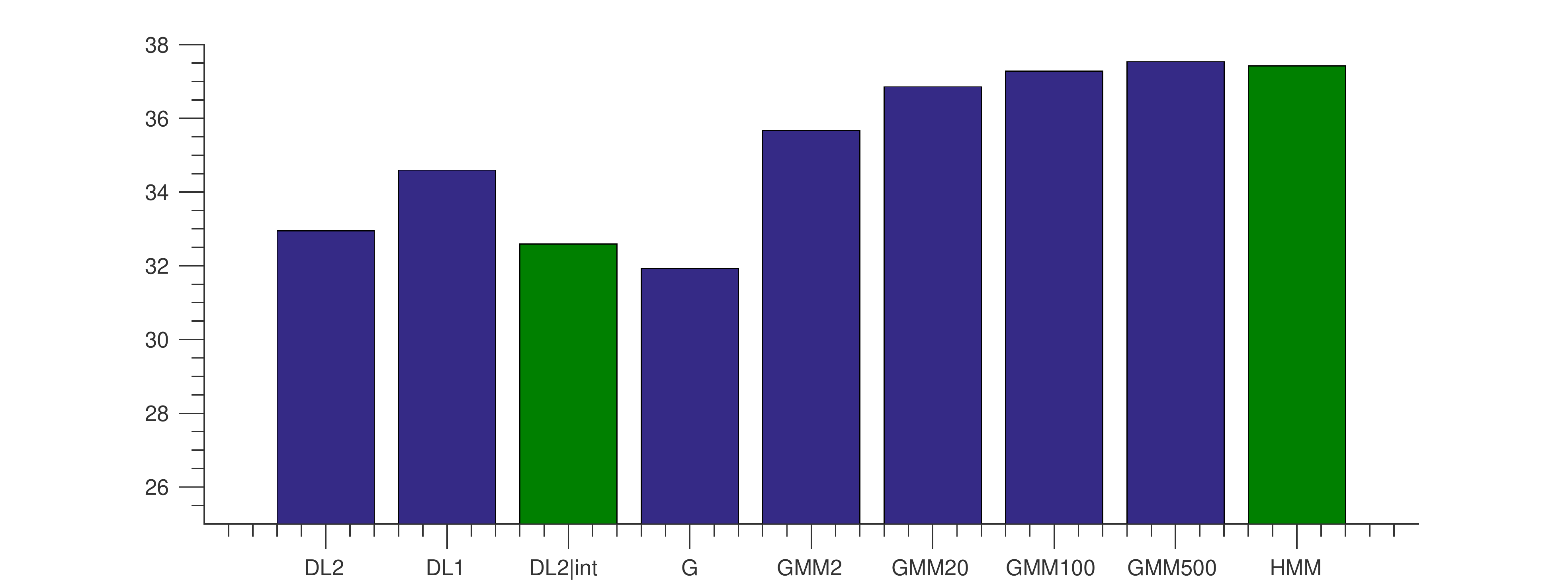}\end{tabular} \\
\end{tabular}
\caption{Patch denoising with different noise levels (average PSNR in dB). GMMs with enough components outperform all other models. Conditioning on the intensity does not lead to a significant improvement.}
\label{fig:patchDenoise}
\end{figure}
Figure \ref{fig:patchDenoise} shows the PSNR of BLS patch denoising using white Gaussian noise with 2 different standard deviations. 
Once again we cannot perform BLS inference using DL1 in closed form, instead we perform maximum a-posteriori (MAP) inference. 
We see that DL1 outperforms DL2 even though it is used with MAP inference which is sub-optimal. Figure \ref{fig:patchDenoise} also shows that conditioning on the intensity does not lead to a significant improvement in patch denoising.

\begin{figure}
  \begin{tabular}{cc}
  &\includegraphics[trim={0 0 0 0},clip,width=0.8\textwidth]{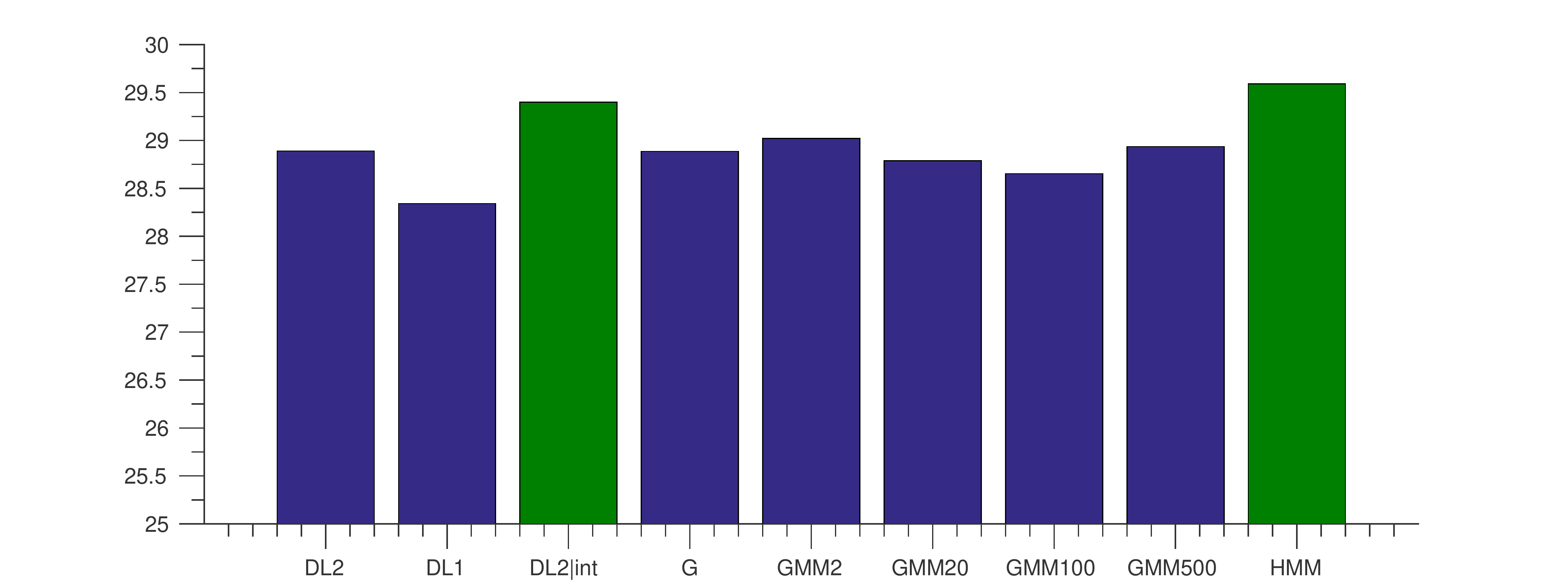}\\[0.3cm]
    \begin{tabular}{c}
      \tiny{intensity}\\[0.05cm]
      \tiny{disparity}\\[0.05cm]
      \tiny{occluded}\\[0.05cm]
      \tiny{GMM500}\\[0.05cm]
      \tiny{DL2$\vert$int}\\[0.05cm]
      \tiny{HMM}
    \end{tabular} &
    \begin{tabular}{c} \includegraphics[trim={0 0 0cm 0},clip, width=0.75\textwidth]{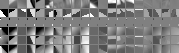}
    \end{tabular}
  \end{tabular}
\caption{Patch inpainting: average PSNR in dB (top) and examples of restored patches (bottom). Conditioning on the intensity leads to a significant improvement. The HMM learned model outperforms all other models.}
\label{fig:patchInpainting}
\end{figure}

In figure \ref{fig:patchInpainting}, we show the results of patch inpainting where most of the patch is hidden and only 4 pixels in 2 corners are visible. 
This is equivalent to denoising with a noise model of very large variance in the hidden pixels. Here we see that conditioning on the intensity does lead to a significant improvement in the PSNR.
The images on the bottom show some examples of the intensity, disparity, occluded disparity and restored disparity patches. We see that DL2$\vert$int does very well when there is a strong match between the disparity and intensity.

\section{Learning density models}
A natural question at this point is if we can use the available training set to learn better models of the disparity. Following the success in learning Gaussian Mixture Models (GMM) for natural image priors~\cite{zoran2011learning} and optical flow~\cite{rosenbaum2013learning}, we train a GMM model with a fixed mean and full covariance matrices over patches of $8 \times 8$ pixels:
\begin{equation}
  Pr(d) = \sum_{k=1}^K \pi(k) \frac{1}{Z_k} e^{-\frac{1}{2}(d-d_0)\top \Sigma_k^{-1} (d-d_0)}
\end{equation}
We use the expectation maximization (EM) algorithm for training. The GMM has many parameters so we emphasize that the different evaluations are performed on a held-out test-set that was not used for training. 

Figure \ref{fig:logl} shows the log-likelihood on the test-set for a single Gaussian (G) and GMMs with a different number of components along with the hand-crafted models. We see that the Gaussian has a very similar log-likelihood to DL2, and that GMMs with enough components outperform other models.

Figure \ref{fig:gsamps} shows patches that were randomly generated using the single Gaussian and the different GMMs. We see that (1) G has a very similar behavior as DL2, (2) GMM2 has mostly very flat patches and occasionally a noisy one, and (3) GMM100 and GMM500 capture the property that whenever a patch is not flat, it is likely to contain an edge with a certain orientation and translation.
The patches generated by GMM500 appear very similar to the ground truth patches.

Figure \ref{fig:patchDenoise} and Figure \ref{fig:patchInpainting} show that also in terms of patch restoration, a GMM with enough components outperforms any independent model (which does not depend on intensity), however even a GMM with 500 components is outperformed by DL2$\vert$int when the dependence on intensity is critical, like in inpainting. The bottom image in figure \ref{fig:patchInpainting} shows that it is hopeless to expect an independent model to recover some of the patches given only 4 visible pixels. In the next section we describe a learned conditional model, but first we elaborate on the GMM.

\begin{figure}
\centering
\renewcommand{\tabcolsep}{1.5pt}
\begin{tabular}{lllc}
 &  & \small{leading eigenvectors} & \small{randomly generated samples} \\
\begin{tabular}{c} \tiny{G}\end{tabular} & \begin{tabular}{c}\tiny{$\pi=1$}\end{tabular} & 
\begin{tabular}{c} \includegraphics[width=0.17\textwidth]{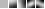} \end{tabular}&
\begin{tabular}{c} \includegraphics[width=0.5\textwidth]{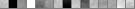} \end{tabular}\\
\begin{tabular}{c} \tiny{GMM2}\end{tabular} & \begin{tabular}{c} \tiny{$\pi=0.82$} \\ \tiny{$\pi=0.18$} \end{tabular} & 
\begin{tabular}{c} \includegraphics[width=0.17\textwidth]{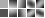} \end{tabular}&
\begin{tabular}{c} \includegraphics[width=0.5\textwidth]{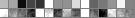} \end{tabular}\\

\begin{tabular}{c} \tiny{GMM100}\end{tabular} & \begin{tabular}{c} \tiny{$\pi=0.5$} \\ \tiny{$\pi=0.19$} \end{tabular} & 
\begin{tabular}{c} \includegraphics[width=0.17\textwidth]{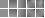} \end{tabular}&
\begin{tabular}{c} \includegraphics[width=0.5\textwidth]{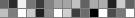} \end{tabular}\\

 & \begin{tabular}{c} \tiny{$\pi \approx 10^{-3}$}\end{tabular} & 
\begin{tabular}{c} \includegraphics[width=0.17\textwidth]{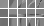} \end{tabular}&
\begin{tabular}{c} \includegraphics[width=0.5\textwidth]{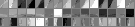} \end{tabular}
\end{tabular}
\caption{Leading eigen-vectors and generated samples from the single Gaussian, from the 2 components of GMM2 and from some of the components of GMM100. As more components are used, the GMM learns to explicitly model flat patches and edges with different orientations and translations.}
\label{fig:GMMcomp}
\end{figure}

The GMM is a model with a single discrete hidden variable which is the index of the Gaussian component. This hidden component has a prior distribution which is the mixing-weights. The division of the $64$ dimensional space of disparity patches into different components can be seen as a way to concentrate the density around different subspaces. Figure \ref{fig:GMMcomp} shows how the space is divided as we train GMMs with more components: The first line shows what a single Gaussian learns. On the left we show the leading 5 eignevectors of the covariance matrix and on the right we show patches generated from the Gaussian. As we've seen before the behavior is very similar to DL2 which is also a Gaussian model. The second and third line show the leading eignevectors of the covariance and generated samples from the 2 components of GMM2. We see that there is an explicit division between very flat patches that occur in probability $0.82$ (as shown by the mixing weight on the left), and noisy patches in probability $0.18$. When we train GMMs with more components we see the explicit assignment of every component to either a flat patch or to a patch with an edge in a certain orientation and translation. We show here only a subset of 5 components. 

\subsection{Learning the dependence on intensity}
In order to capture a possible dependence on intensity, we train on top of the GMM500 another model called an HMM as was done in \cite{rosenbaum2013learning}. The HMM is built of 2 GMMs: the first is a GMM over the intensity like in \cite{zoran2011learning}, and the second one is a GMM over the disparity but instead of having independent mixing weights (i.e. a prior on the component), the disparity component depends on the intensity component through a transition matrix. The HMM is equivalent to having a GMM model over the disparity with mixing weights that change according to the intensity. Since the intensity GMM also assigns different components to different orientations and translations of edges, this allows the occurrence of intensity edges to give a higher prior for disparity edges in the same orientation and translation.

Looking at the generated samples in figure \ref{fig:gsampsGivenInt} we see that this is exactly what the HMM does. Given an intensity edge, disparity edge components with similar orientation and translation become more likely. Note that this intensity dependent prior is `soft' and allows also flat patches and edges in very different orientation and translation to occur but in a lower probability. 
If we compare the HMM samples to the DL2$\vert$int samples we see that DL2$\vert$int has the advantage of being able to exactly match the intensity edge however it lacks the power of the HMM to model the non-negligible probability of similar orientations and translation of edges as the ground truth data also exhibits in figure \ref{fig:intVsDisp}.

In terms of log-likelihood and patch restoration, the HMM model is superior to all other models in all the different evaluations. It has similar results to the GMM500 in log-likelihood (figure \ref{fig:logl}), and patch denoising (figure \ref{fig:patchDenoise}), and outperforms it when the dependence on intensity is needed for inpatining (figure \ref{fig:patchInpainting}). For inpainting it also outperforms the hand-crafted conditional model DL2$\vert$int.

\section{Disparity estimation in full images}
Given the superior performance on patches, we would like to use the learned models to perform disparity estimation in full images. 
As long as the degredation in disparity is local and contains noise and small holes, a simple approach is to perform patch restoration on all overlapping patches in the image and average the results over overlapping pixels. However, when there are big holes as in the dataset used in \cite{Lu2014depth}, global inference is needed.
While the hand-crafted models DL2, DL1 and DL2$\vert$int can be extended to a full image model, for the GMMs it is not feasible. 
The reason is that extending a mixture model over patches to an image with thousands or milions of patches would require to go over all the combination of mixture components. 
Moreover, since the model was learned over patches it cannot capture the depndence between neighboring (or even overlapping) patches. One option is to treat all patches as independent and perform global MAP inference. This is shown to work succefully in the EPLL framework of \cite{zoran2011learning}. Another implementation of global MAP inference can be done using the EM-MAP method~\cite{levi2009using}. This is performed iteratively by building a sparse inverse covariance matrix over the whole image and inverting it in each iteration. 

\begin{figure}
\begin{tabular}{cc}
 $\sigma=5/255$ &  $\sigma=15/255$ \\
\includegraphics[width=0.5\textwidth]{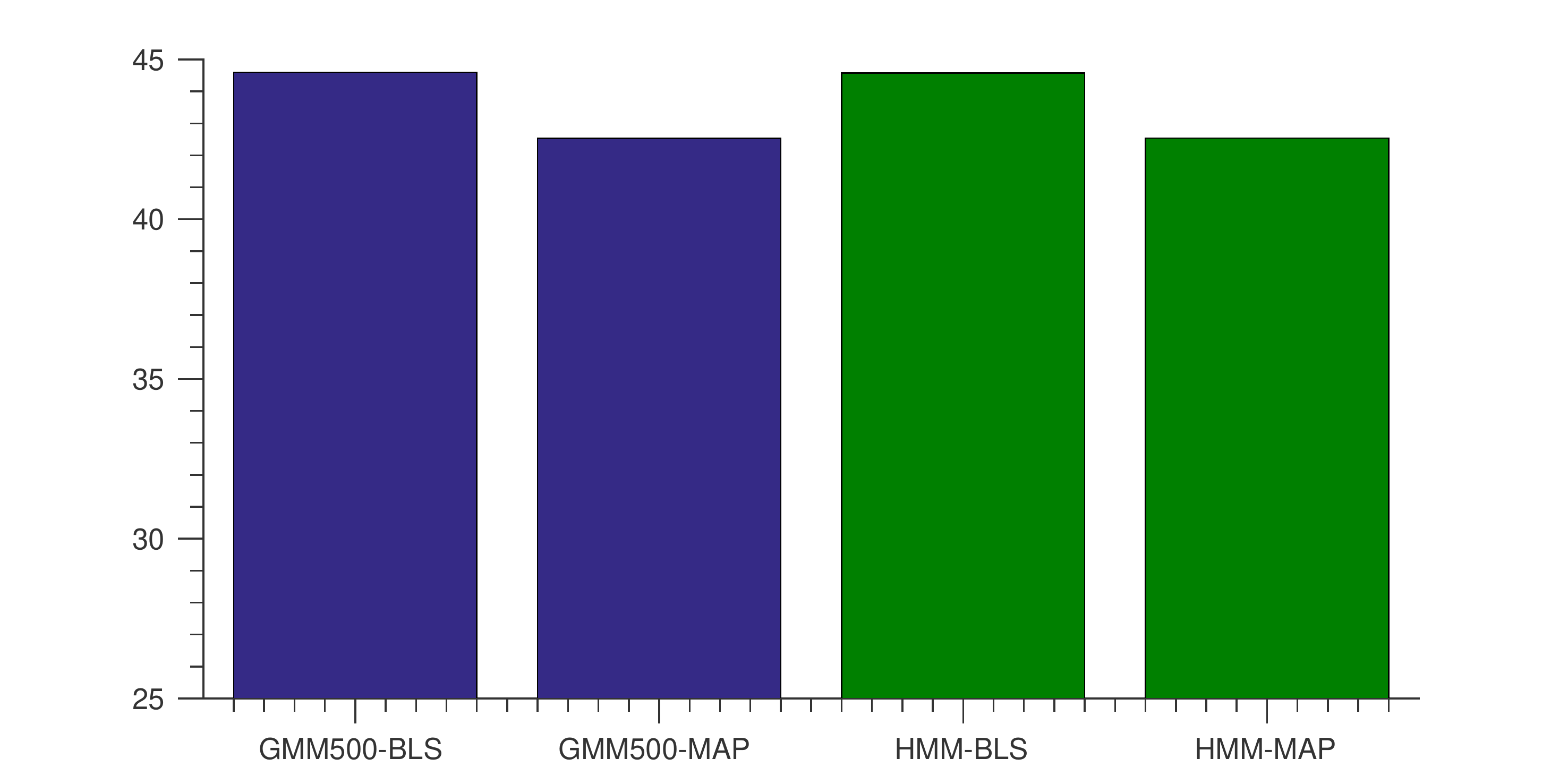}&
\includegraphics[width=0.5\textwidth]{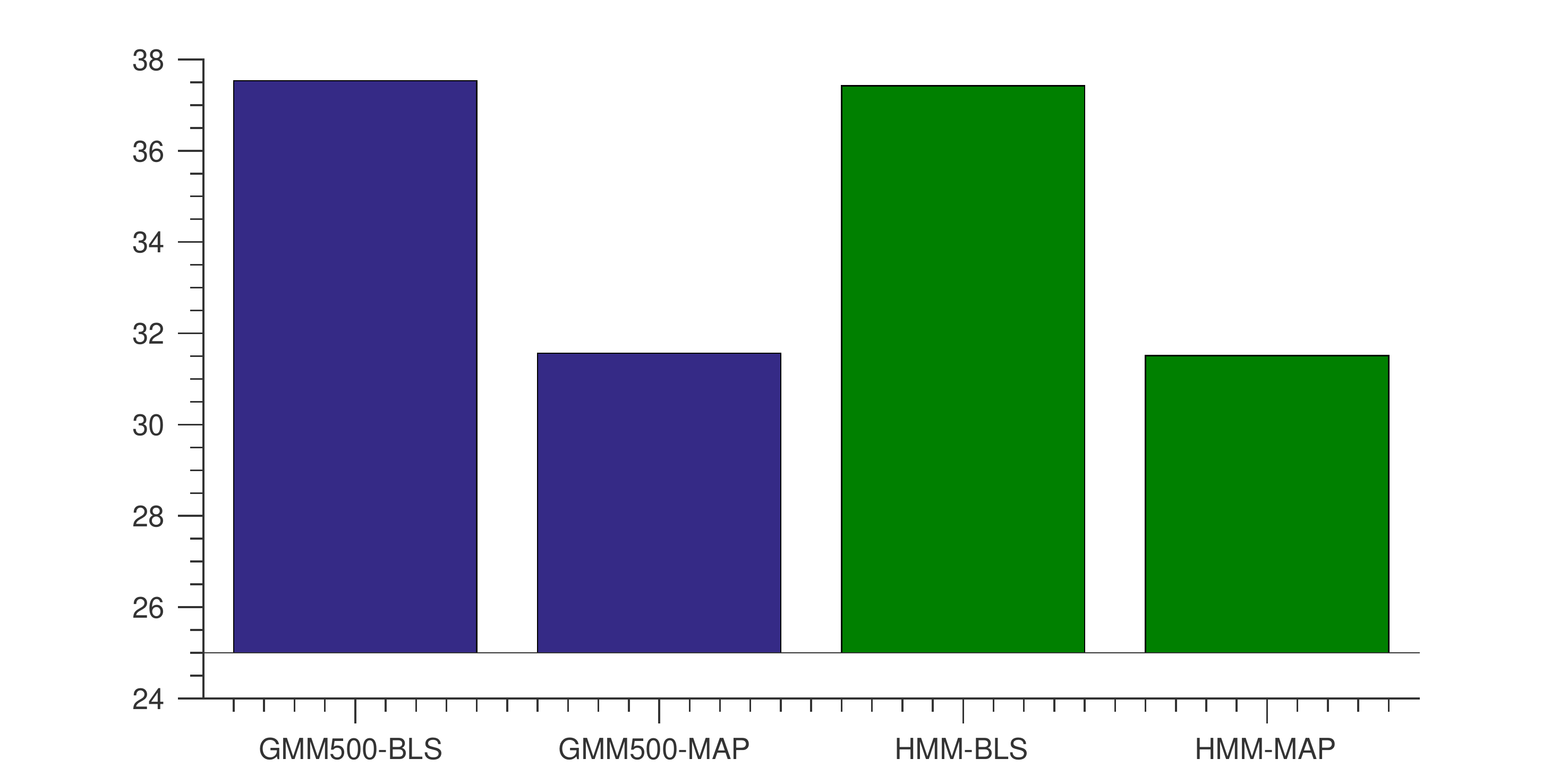}\\[0.2cm]
inpainting&\\
\includegraphics[width=0.5\textwidth]{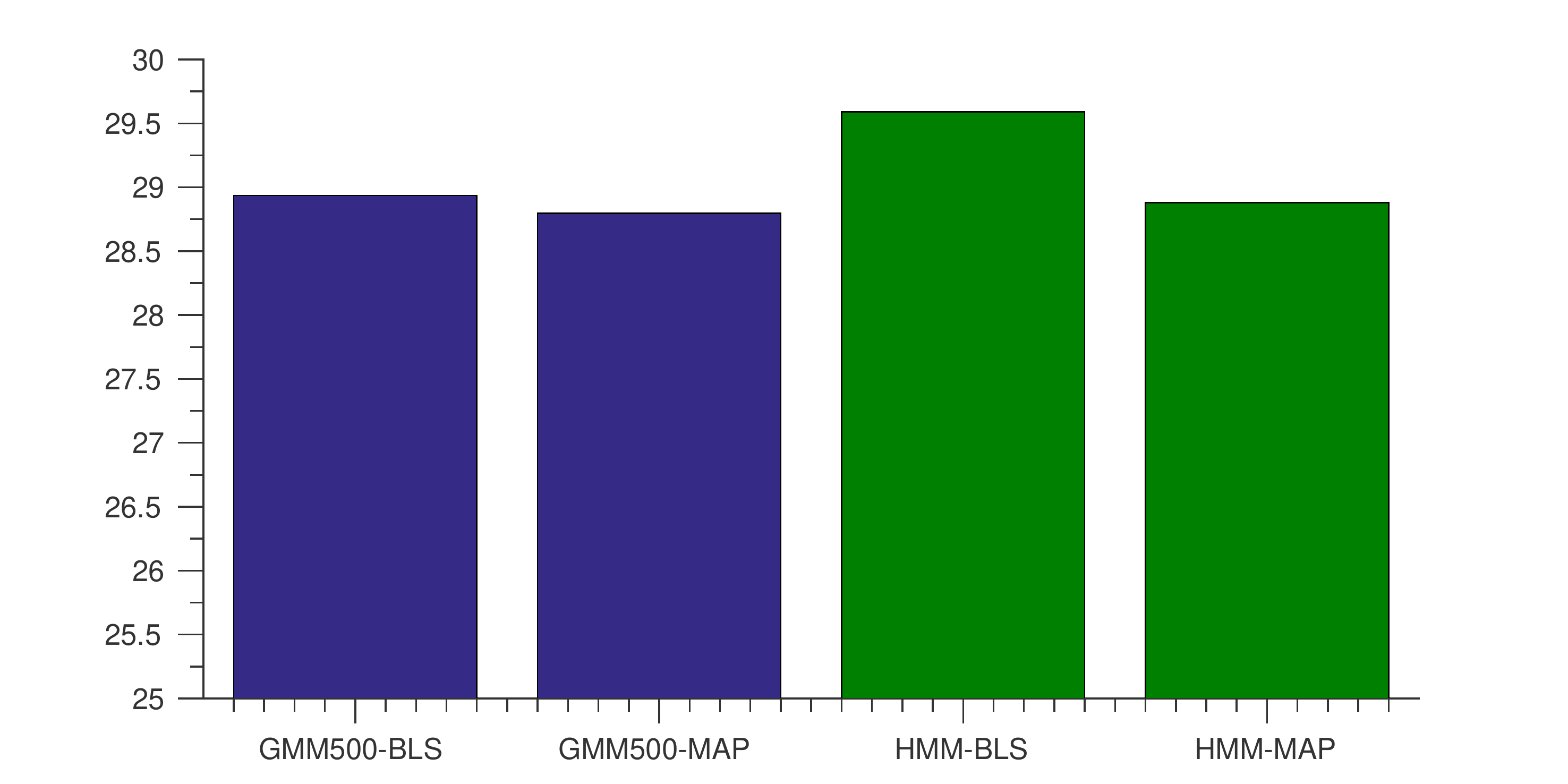}
\end{tabular}
\caption{BLS vs MAP inference for the GMM500 and HMM models. MAP inference is inferior in both patch denoising and inpainting.}
\label{fig:patchMAP}
\end{figure}

However, one drawback of these methods is that even if the optimization succeeds,  the MAP solution is not guarenteed to have good performance even for good density models.
In fact, if we evaluate the result of MAP inference over patches we see that it is significantly inferior to BLS inference (see~\cite{schmidt2010generative} for a similar result in image restoration). Figure \ref{fig:patchMAP} shows that the performance drops for both denoising and inpaintining once we turn to MAP inference. For inpainting we see that the gap between the HMM and the GMM, which was due to the dependence on intensity, disappears. The performance of HMM-MAP is also worse than the performance of DL2$\vert$int (for which MAP and BLS inference are the same).

Therefore, in order to restore a given disparity image that contains noise and holes, we do the following 2 steps:
\begin{enumerate}
\item We perform BLS inference using the HMM over all overlapping patches in the image and average the results over overlapping pixels.
\item Using the resulting image, we perform global BLS inference on the large holes using the DL2$\vert$int model.
\end{enumerate}

We run this procedure on the online availabe dataset used by Lu et al~\cite{Lu2014depth} which consists of 30 images from Middlebury~\cite{baker2011database} and 9 images from the RGBZ dataset~\cite{CGF:CGF3003}. The noisy intensity image is denoised using EPLL~\cite{zoran2011learning}.  We compare our proposed method (HMM+DL2$\vert$int) to only using global inference with DL2$\vert$int and to the methods that were compared in \cite{Lu2014depth}. These methods include the Joint Bilateral Filter (JBF)~\cite{richardt2012coherent} and the LRC method of Lu et al. that assumes that concatenated vectors of disparity patches and corresponding color patches lie in a low rank subspace. Our proposed method acheives an improvement in average PSNR of almost 1dB over the state-of-the-art results of LRC. 

\begin{figure}
\begin{tabular}{ccc}
noisy intensity & GT disparity & noisy disparity \\
\includegraphics[width=0.32\textwidth]{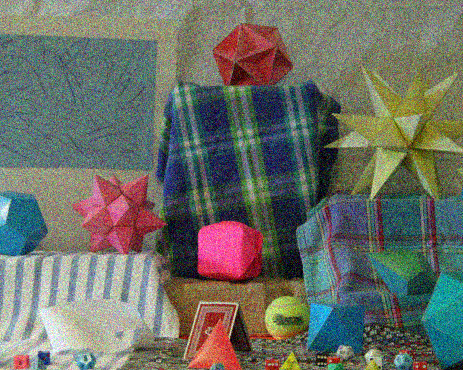}&
\includegraphics[width=0.32\textwidth]{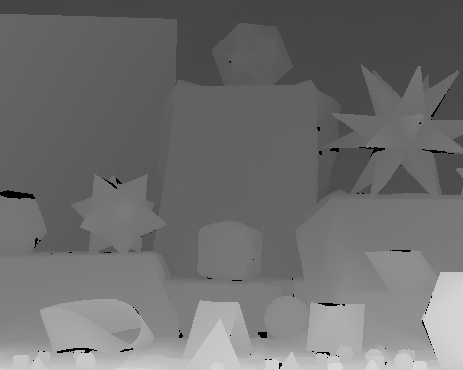}&
\includegraphics[width=0.32\textwidth]{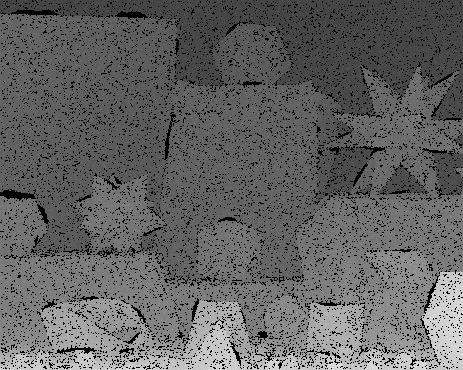}\\[0.3cm]
LRC (41.38) &  DL2$\vert$int (37.89) & HMM+DL2$\vert$int (42.22)\\
\includegraphics[width=0.32\textwidth]{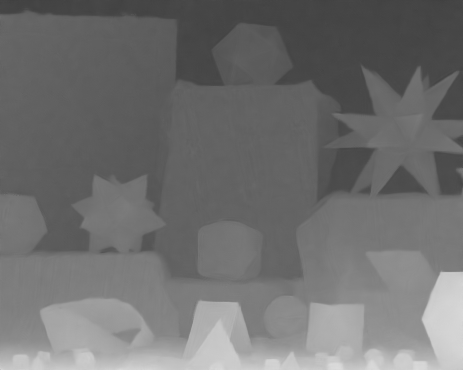}&
\includegraphics[width=0.32\textwidth]{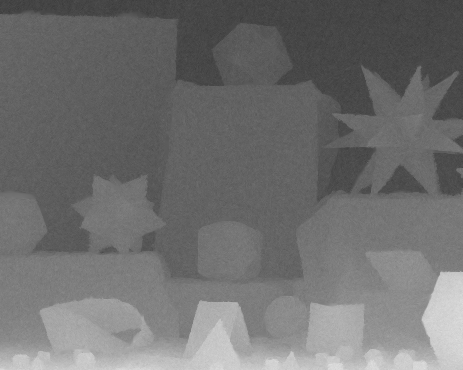}&
\includegraphics[width=0.32\textwidth]{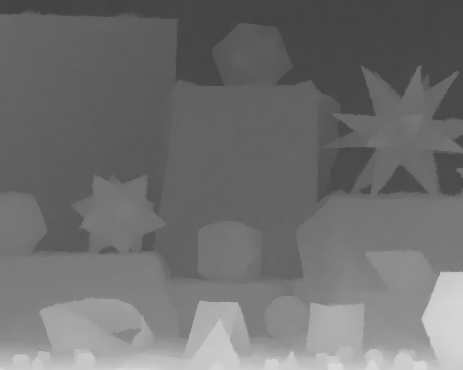}\\[0.5cm]

noisy intensity & GT disparity & noisy disparity \\
\includegraphics[width=0.32\textwidth]{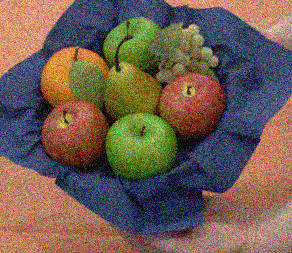}&
\includegraphics[width=0.32\textwidth]{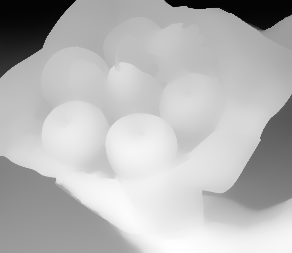}&
\includegraphics[width=0.32\textwidth]{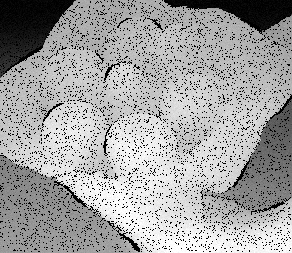}\\[0.3cm]
LRC (36.44) &  DL2$\vert$int (35.88) & HMM+DL2$\vert$int (37.96) \\
\includegraphics[width=0.32\textwidth]{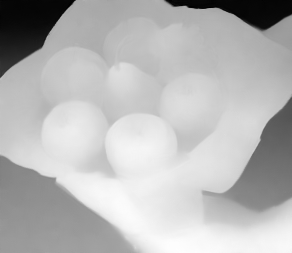}&
\includegraphics[width=0.32\textwidth]{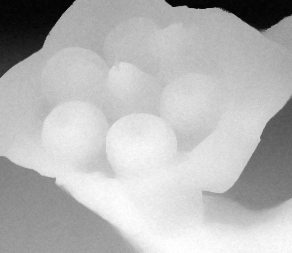}&
\includegraphics[width=0.32\textwidth]{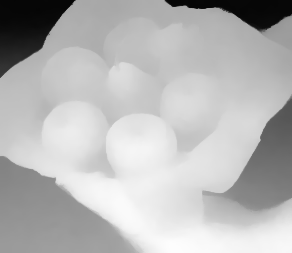}\\
\end{tabular}
\caption{Examples of disparity images enhanced with LRC, DL2$\vert$int and HMM+DL2$\vert$int. PSNR values are in dB.}
\label{fig:results}
\end{figure}

Table \ref{tab:results} shows the average PSNR of HMM+DL2$\vert$int, DL2$\vert$int, and the different methods that were compared in \cite{Lu2014depth}.
Figure \ref{fig:results} shows examples of our results compared to LRC and using only DL2$\vert$int. 
We emphasize that even though the models were trained on the synthetic data of Sintel, we acheive a significant improvement on the Middlebury+RGBZ dataset of real images. 

\begin{table}
\centering
\begin{tabular}{c|c|c|c|c|c|c|c}
HMM+DL2int ~& ~DL2int~ & ~LRC~ & ~JBF~ & ~NLM~ & ~SGF~ & ~SHF~ & ~GIF~ \\
\hline
{\bf 40.2} & 36.7 & 39.3 & 37.9 & 37.2 & 33.9 & 36.5 & 37.0 \\
\end{tabular}  
~\\[0.5cm]
\caption{Average PSNR (in dB) of DL2$\vert$int, HMM+DL2$\vert$int and the methods that were compared in \cite{Lu2014depth}.} 
\label{tab:results}
\end{table}

\section{Discussion}
An advantage of using learning based approaches for vision is that we can compare what is learned to assumptions commonly made by Computer Vision researchers. The majority of previous approaches to improving D given RGB used the assumption that depth edges are correlated with intensity edges and assumed very little additional structure on the depth. In this paper we have shown that a generative model that is learned from ground truth RGBD patches indeed finds a correlation between depth edges and intensity edges but this correlation is relatively weak. At the same time, the generative model learns very strong structural constraints on the depth: that depth patches are usually either flat or edges. By using a learned model that combines both the depth structure and the correlation with intensity we were able to outperform the state-of-the-art in improving the quality of the depth channel given RGB. Even though our training was performed on synthetic images, we gained a significant advantage (about 1dB on average) in restoring real images.


\bibliographystyle{splncs}
\bibliography{bib}
\end{document}